# The Difficulties of Learning Logic Programs with Cut

**Francesco Bergadano**                                    BERGADAN@DI.UNITO.IT
*Università di Catania, Dipartimento di Matematica,*
*via Andrea Doria 6, 95100 Catania, Italy*

**Daniele Gunetti**                                        GUNETTI@DI.UNITO.IT
**Umberto Trinchero**                                      TRINCHER@DI.UNITO.IT
*Università di Torino, Dipartimento di Informatica,*
*corso Svizzera 185, 10149 Torino, Italy*

## Abstract

As real logic programmers normally use cut (!), an effective learning procedure for logic programs should be able to deal with it. Because the cut predicate has only a procedural meaning, clauses containing cut cannot be learned using an extensional evaluation method, as is done in most learning systems. On the other hand, searching a space of possible programs (instead of a space of independent clauses) is unfeasible. An alternative solution is to generate first a candidate base program which covers the positive examples, and then make it consistent by inserting cut where appropriate. The problem of learning programs with cut has not been investigated before and this seems to be a natural and reasonable approach. We generalize this scheme and investigate the difficulties that arise. Some of the major shortcomings are actually caused, in general, by the need for intensional evaluation. As a conclusion, the analysis of this paper suggests, on precise and technical grounds, that learning cut is difficult, and current induction techniques should probably be restricted to purely declarative logic languages.

## 1. Introduction

Much recent research in AI and Machine Learning is addressing the problem of learning relations from examples, especially under the title of Inductive Logic Programming (Muggleton, 1991). One goal of this line of research, although certainly not the only one, is the inductive synthesis of logic programs. More generally, we are interested in the construction of program development tools based on Machine Learning techniques. Such techniques now include efficient algorithms for the induction of logical descriptions of recursive relations. However, real logic programs contain features that are not purely logical, most notably the cut (!) predicate. The problem of learning programs with cut has not been studied before in Inductive Logic Programming, and this paper analyzes the difficulties involved.

### 1.1 Why Learn Programs with Cut?

There are two main motivations for learning logic programs with cut:

1. ILP should provide practical tools for developing logic programs, in the context of some general program development methodology (e.g., (Bergadano, 1993b)); as real size logic programs normally contain cut, learning cut will be important for creating an integrated Software Engineering framework.

2. Extensive use of cut can make programs sensibly shorter, and the difficulty of learning a given logic program is very much related to its length.

For both of these objectives, we need not only cuts that make the programs more efficient without changing their input-output behavior ("green cuts"), but also cuts that eliminate some possible computed results ("red cuts"). Red cuts are sometimes considered bad programming style, but are often useful. Moreover, only the red cuts are effective in making programs shorter. Green cuts are also important, and less controversial. Once a correct program has been inferred via inductive methods, it could be made more efficient through the insertion of green cuts, either manually or by means of automated program transformation techniques (Lau & Clement, 1993).

## 1.2 Why Standard Approaches Cannot be Used?

Most Machine Learning algorithms generate rules or clauses one at a time and independently of each other: if a rule is useful (it covers some positive example) and correct (it does not cover any negative example), then it is added to the description or program which is being generated, until all positive examples have been covered. This means that we are searching a space of possible clauses, without backtracking. This is obviously a great advantage, as programs are sets of clauses, and therefore the space of possible programs is exponentially larger.

The one principle which allows this simplification of the problem is the *extensional* evaluation of possible clauses, used to determine whether a clause C *covers* an example $e$. The fact that a clause C covers an example $e$ is then used as an approximation of the fact that a logic program containing C derives $e$. Consider, for instance, the clause C = "p(X,Y) ← $\alpha$", and suppose the example $e$ is p(a,b). In order to see whether C covers $e$, the extensionality principle makes us evaluate any literal in $\alpha$ as true if and only if it matches some given positive example. For instance, if $\alpha$ = q(X,Z) ∧ p(Z,Y), then the example p(a,b) is *extensionally covered* iff there is a ground term c such that q(a,c) and p(c,b) are given as positive examples. In particular, in order to obtain the truth value of p(c,b), we will not need to call other clauses that were learned previously. For this reason, determining whether C covers $e$ only depends on C and on the positive examples. Therefore, the learning system will decide whether to accept C as part of the final program P independently of the other clauses P will contain.

The extensionality principle is found in Foil (Quinlan, 1990) and its derivatives, but is also used in bottom-up methods such as Golem (Muggleton & Feng, 1990). Shapiro's MIS system (Shapiro, 1983) uses it when refining clauses, although it does not when backtracing inconsistencies. We have also used an extensional evaluation of clauses in the FILP system (Bergadano & Gunetti, 1993).

When learning programs with cut, clauses are no longer independent and their stand-alone extensional evaluation is meaningless. When a cut predicate is evaluated, other possible clauses for proving the same goal will be ignored. This changes the meaning of these other clauses. Even if a clause extensionally covers some example $e$, it may be the case that the final program does not derive $e$, because some derivation paths have been eliminated by the evaluation of a cut predicate.

However, an exhaustive search in a space of programs is prohibitive. Learning methods, even if based on extensionality, are often considered inefficient if sufficient prior information is not available; searching for *sets* of clauses will be exponentially worse. This would amount to a brute-force enumeration of all possible logic programs containing cut, until a program that is consistent with the given examples is found.

## 1.3 Is there an Alternative Method?

Cut will only eliminate some computed results, i.e., after adding cut to some program, it may be the case that some example is no longer derived. This observation suggests a general learning strategy: a base program P is induced with standard techniques, given the positive and maybe some of the negative examples, then the remaining negative examples are ruled out by inserting cut in some clause of P. Obviously, after inserting cut, we must make sure that the positive examples may still be derived.

Given the present technology and the discussion above, this seems to be the only viable path to a possible solution. Using standard techniques, the base program P would be generated one clause at a time, so that the positive examples are extensionally covered. However, we think this view is too restrictive, as there are programs which derive all given positive examples, although they do not cover them extensionally (Bergadano, 1993a; DeRaedt, Lavrac, & Dzeroski, 1993). More generally, we consider traces of the positive examples:

**Definition 1** *Given a hypothesis space S of possible clauses, and an example e such that S ⊢ e, the set of clauses T⊆S which is used during the derivation of e is called a trace for e.*

We will use as a candidate base program P any subset of S which is the union of some traces for the positive examples. If P⊆S extensionally covers the positive examples, then it will also be the union of such traces, but the converse is not always true. After a candidate program has been generated, an attempt is made to insert cuts so that the negative examples are not derived. If this is successful, we have a solution, otherwise, we backtrack to another candidate base program. We will analyze the many problems inherent in learning cut with this class of trace-based learning methods, but, as we discuss later (Section 4), the same problems need to be faced in the more restrictive framework of extensional evaluation. In other words, even if we choose to learn the base program P extensionally, and then we try to make it consistent by using cut, the same computational problems would still arise. The main difference is that standard approaches based on extensionality do not allow for backtracking and do not guarantee that a correct solution is found (Bergadano, 1993a).

As far as computational complexity is concerned, trace-based methods have a complexity standing between the search in a space of independent clauses (for the extensional methods) and the exhaustive search in a space of possible programs. We need the following:

**Definition 2** *Given a hypothesis space S, the* depth *of an example e is the maximum number of clauses in S successfully used in the derivation of e.*

For example, if we are in a list processing domain, and S only contains recursive calls of the type "P([H|T]) :- ..., P(T), ..." then the depth of an example P(L) is the length of L. For practical program induction tasks, it is often the case that the depth of an example is

related to its complexity, and not to the hypothesis space S. If $d$ is the maximum depth for the given $m$ positive examples, then the complexity of trace-based methods is of the order of $|S|^{md}$, while extensional methods will just enumerate possible clauses with a complexity which is linear in $|S|$, and enumerating all possible programs is exponential in $|S|$.

## 2. A Simple Induction Procedure

The trace-based induction procedure we analyze here takes as input a finite set of clauses S and a set of positive and negative examples E+ and E- and tries to find a subset T of S such that T derives all the positive examples and none of the negative examples. For every positive example e+ $\in$ E+, we assume that S is large enough to derive it. Moreover, we assume that all clauses in S are flattened[1]. If this is not the case, clauses are flattened as a preprocessing step.

We consider one possible proof for S $\vdash$ e+, and we build an intermediate program T $\subseteq$ S containing a trace of the derivation. The same is done for the other positive examples, and the corresponding traces T are merged. Every time T is updated, it is checked against the negative examples. If some of them are derived from T, cut (!) is inserted in the antecedents of the clauses in T, so that a consistent program is found, if it exists. If this is not the case, the procedure backtracks to a different proof for S $\vdash$ e+. The algorithm can be informally described as follows:

input: a set of clauses S
        a set of positive examples E+
        a set of negative examples E-
S := flatten(S)
T $\leftarrow \emptyset$
For each positive example e+ $\in$ E+
        find T1 $\subseteq$ S such that T1 $\vdash_{SLD}$ e+ (backtracking point 1)
        T $\leftarrow$ T $\cup$ T1
        if T derives some negative example e- then trycut(T,e-)
        if trycut(T,e-) fails then backtrack
output the clauses listed in T

trycut(T,e-):
insert ! somewhere in T (backtracking point 2) so that
        1. all previously covered positive examples are still derived from T, and
        2. T $\not\vdash_{SLD}$ e-

The complexity of adding cut somewhere in the trace T, so that the negative example e- is no longer derived, obviously only depends on the size of T. But this size depends on the depth of the positive examples, not on the size of the hypothesis space S. Although more clever ways of doing this can be devised, based on the particular example e-, we propose a simple enumerative technique in the implementation described in the Appendix.

---

1. A clause is *flattened* if it does not contain any functional symbol. Given an unflattened clause, it is alway possible to flatten it (by turning functions into new predicates with an additional argument representing the result of the function) and vice versa (Rouveirol, in press).

## 3. Example: Simplifying a List

In this section we show an example of the use of the induction procedure to learn the logic program "*simplify*". *Simplify* takes as input a list whose members may be lists, and transforms it into a "flattened" list of single members, containing no repetitions and no lists as members. This program appears as exercise number 25 in (Coelho & Cotta, 1988), is composed of nine clauses (plus the clauses for *append* and *member*); six of them are recursive, one is doubly-recursive and cut is extensively used. Even if *simplify* is a not a very complex logic program, it is more complex than usual ILP test cases. For instance, the *quicksort* and *partition* program, which is very often used, is composed of only five clauses (plus those for *append*), and three of them are recursive. Moreover, note that the conciseness of *simplify* is essentially due to the extensive use of cut. Without cut, this program would be much longer. In general, the longer a logic program, the more difficult to learn it.

As a consequence, we start with a relatively strong bias; suppose that the following hypothesis space of N=8449 possible clauses is defined by the user:

- The clause "simplify(L,NL) :- flatten(L,L1), remove(L1,NL)."

- All clauses whose head is "flatten(X,L)" and whose body is composed of a conjunction of any of the following literals:

  head(X,H), tail(X,L1), equal(X,[L1,T]), null(T), null(H), null(L1), equal(X,[L1]), flatten(H,X1), flatten(L1,X2), append(X1,X2,L), assign(X1,L), assign(X2,L), list(X,L).

- All clauses whose head is "remove(IL,OL)" and whose body is composed of a conjunction of any of the following literals:

  cons(X,N,OL), null(IL), assign([],OL), head(IL,X), tail(IL,L), member(X,L), remove(L,OL), remove(L,N).

- The correct clauses for *null*, *head*, *tail*, *equal*, *assign*, *member*, *append* are given:

  null([]).
  head([H|_],H).
  tail([_|T],T).
  equal(X,X).
  assign(X,X).
  member(X,[X|_]).
  member(X,[_|T]) :- member(X,T).

```
      append([],Z,Z).
      append([H|X],Y,[H|Z]) :- append(X,Y,Z).
```

By using various kinds of constraints, the initial number of clauses can be strongly reduced. Possible constraints are the following:

- Once an output is produced it must not be instantiated again. This means that any variable cannot occur as output in the antecedent more than once.

- Inputs must be used: all input variables in the head of a clause must also occur in its antecedent.

- Some conjunctions of literals are ruled out because they can never be true, e.g. null(IL)∧head(IL,X).

By applying various combination of these constraints it is possible to strongly restrict the initial hypothesis space, which is then given in input to the learning procedure. The set of positive and negative examples used in the learning task is:

```
simplify_pos([[[],[b,a,a]],[]],[b,a]).   remove_pos([a,a],[a]).
(simplify_neg([[[],[b,a,a]],[]],X),not equal(X,[b,a])).
simplify_neg([[a,b,a],[]],[a,[b,a]]).    remove_neg([a,a],[a,a]).
```

Note that we define some negative examples of *simplify* to be all the examples with the same input of a given positive example and a different output, for instance simplify_neg([[[],[b,a,a]],[]],[a,b]). Obviously, it is also possible to give negative examples as normal ground literals. The learning procedure outputs the program for *simplify* reported below, which turns out to be substantially equivalent to the one described in (Coelho & Cotta, 1988) (we have kept clauses unflattened).

```
simplify(L,NL) :- flatten(L,L1), remove(L1,NL).

flatten(X,L) :- equal(X,[L1,T]), null(T), !, flatten(L1,X2), assign(X2,L).
flatten(X,L) :- head(X,H), tail(X,L1), null(H), !, flatten(L1,X2), assign(X2,L).
flatten(X,L) :- equal(X,[L1]), !, flatten(L1,X2), assign(X2,L).
flatten(X,L) :- head(X,H), tail(X,L1), !,
                     flatten(H,X1), !, flatten(L1,X2), append(X1,X2,L).
flatten(X,L) :- list(X,L).

remove(IL,OL) :- head(IL,X), tail(IL,L), member(X,L), !, remove(L,OL).
remove(IL,OL) :- head(IL,X), tail(IL,L), remove(L,N), cons(X,N,OL).
remove(IL,OL) :- null(IL), assign([],OL).
```

The learning task takes about 44 seconds on our implementation. However, This is obtained at some special conditions, which are thoroughly discussed in the next sections:

- All the constraints listed above are applied, so that the final hypothesis space is reduced to less than one hundred clauses.

- Clauses in the hypothesis space are generated in the correct order, as they must appear in the final program. Moreover, literals in each clause are in the correct position. This is important, since in a logic program with cut the relative position of clauses and literals is significant. As a consequence, we can learn *simplify* without having to test for different clause and literal orderings (see subsections 4.2 and 4.5).

- We tell the learning procedure to use at most two cuts per clause. This seems to be quite an intuitive constraint since, in fact, many classical logic programs have no more than one cut per clause (see subsections 4.1 and 5.4).

## 4. Problems

Experiments with the above induction procedure have shown that many problems arise when learning logic programs containing cut. In the following, we analyze these problems, and this is a major contribution of the present paper. As cut cannot be evaluated extensionally, this analysis is general, and does not depend on the specific induction method adopted. Some possible partial solutions will be discussed in Section 5.

### 4.1 Problem 1: Intensional Evaluation, Backtracking and Cut

The learning procedure of Section 2 is very simple, but it can be inefficient. However, we believe this is common to every intensional method, because clauses cannot be learned independently of one another. As a consequence, backtracking cannot be avoided and this can have some impact on the complexity of the learning process. Moreover, cut must be added to every trace covering negative examples. If no constraints are in force, we can range from only one cut in the whole trace to a cut between each two literals of each clause in the trace. Clearly, the number of possibilities is exponential in the number of literals in the trace. Fortunately, this number is usually much smaller than the size of the hypothesis space, as it depends on the depth of the positive examples.

However, backtracking also has some advantages; in particular, it can be useful to search for alternative solutions. These alternative programs can then be confronted on the basis of any required characteristic, such as simplicity or efficiency. For example, using backtracking we discovered a version of *simplify* equivalent to the one given but without the cut predicate between the two recursive calls of the fourth clause of *flatten*.

### 4.2 Problem 2: Ordering of Clauses in the Trace

In a logic program containing cut, the mutual position of clauses is significant, and a different ordering can lead to a different (perhaps wrong) behavior of the program. For example, the following program for *intersection*:

$c_1$) int(X,S2,Y) :- null(X), null(Y).
$c_2$) int(X,S2,Y) :- head(X,H), tail(X,Tail), member(H,S2), !, int(Tail,S2,S), cons(H,S,Y).
$c_3$) int(X,S2,Y) :- head(X,H), tail(X,Tail), int(Tail,S2,Y).

behaves correctly only if $c_2$ comes before $c_3$. Suppose the hypothesis space given in input to the induction procedure consists of the same three clauses as above, but with $c_3$ before $c_2$. If $\neg int([a],[a],[])$ is given as a negative example, then the learning task fails, because clauses $c_1$ and $c_3$ derive that example.

In other words, learning a program containing cut means not only to learn a set of clauses, but also a specific ordering for those clauses. In terms of our induction procedure this means that for every trace T covering some negative example, we must check not only every position for inserting cuts, but also every possible clause ordering in the trace. This "generate and test" behavior is not difficult to implement, but it can dramatically decrease the performance of the learning task. In the worst case all possible permutations must be generated and checked, and this requires a time proportional to $(md)!$ for a trace of $md$ clauses[2].

The necessity to test for different permutations of clauses in a trace is a primary source of inefficiency when learning programs with cut, and probably the most difficult problem to solve.

## 4.3 Problem 3: Kinds of Given Examples

Our induction procedure is only able to learn programs which are traces, i.e. where every clause in the program is used to derive at least one positive example. When learning definite clauses, this is not a problem, because derivation is monotone, and for every program P, complete and consistent w.r.t. the given examples, there is a program $P' \subseteq P$ which is also complete and consistent and is a trace[3]. On the other hand, when learning clauses containing cut, it may happen that the only complete and consistent program(s) in the hypothesis space is neither a trace, nor contains it as a subset. This is because derivation is no longer monotone and it can be the case that a negative example is derived by a set of clauses, but not by a superset of them, as in the following simple example:

S = {sum(A,B,C) :- A>0, !, M is A-1, sum(M,B,N), C is N+1.
        sum(A,B,C) :- C is B.}
sum_pos(0,2,2), sum_neg(2,2,2).

The two clauses in the hypothesis space represent a complete and consistent program for the given examples, but our procedure is unable to learn it. Observe that the negative example is derived by the second clause, which is a trace for the positive example, but not by the first and the second together.

This problem can be avoided if we require that, for every negative example, a corresponding positive example with the same input be given (in the above case, the example required is sum_pos(2,2,4)). In this way, if a complete program exists in the hypothesis space, then it is also a trace, and can be learned. Then it can be made consistent using cut, in order to rule out the derivation of negative examples. The constraint on positive and negative examples seems to be quite intuitive. In fact, when writing a program, a

---

2. it must be noted that if we are learning programs for two different predicates, of $j$ and $k$ clauses respectively (that is, $md = j+k$), then we have to consider not $(j+k)!$ different programs, but only $j!+k!$. We can do better if, inside a program, it is known that non-recursive clauses have a fixed position, and can be put before or after all the recursive clauses.

3. a learned program P is *complete* if it derives all the given positive examples, and it is *consistent* if it does not derive any of the given negative examples programmer usually thinks in terms of what a program should compute on given inputs, and then tries to avoid wrong computations for those inputs.

## 4.4 Problem 4: Ordering of Given Examples

When learning clauses with cut, even the order of the positive examples may be significant. In the example above, if sum_pos(2,2,4) comes after sum_pos(0,2,2) then the learning task fails to learn a correct program for *sum*, because it cannot find a program consistent w.r.t. the first positive example and the negative one(s).

In general, for a given set of $m$ positive examples this problem can be remedied by testing different example orderings. Again, in the worst case $k!$ different orderings of a set of $k$ positive examples must be checked. Moreover, in some situations a favorable ordering does not exist. Consider the following hypothesis space:

$c_1$) int(X,Y,W) :- head(X,A), tail(X,B), notmember(A,Y), int(B,Y,W).
$c_2$) int(X,Y,W) :- head(X,A), tail(X,B), notmember(A,Y), !, int(B,Y,W).
$c_3$) int(X,Y,Z) :- head(X,A), tail(X,B), int(B,Y,W), cons(A,W,Z).
$c_4$) int(X,Y,Z) :- head(X,A), tail(X,B), !, int(B,Y,W), cons(A,W,Z).
$c_5$) int(X,Y,Z) :- null(Z).

together with the set of examples:

$e_1$) int_pos([a],[b],[ ]).
$e_2$) int_pos([a],[a],[a]).
$e_3$) int_neg([a],[b],[a]).
$e_4$) int_neg([a],[a],[ ]).

Our induction procedure will not be able to find a correct program for any ordering of the two positive examples, even if such a program does exist ($[c_2,c_4,c_5]$). This program is the union of two traces: $[c_2,c_5]$, which covers $e_1$, and $[c_4,c_5]$, which covers $e_2$. Both of these traces are inconsistent, because the first covers $e_4$, and the second covers $e_3$. This problem can be remedied only if all the positive examples are derived before the check against negative examples is done.

However, in that case we have a further loss of efficiency, because some inconsistent traces are discarded only in the end. In other words, we would need to learn a program covering *all* the positive examples, and then make it consistent by using cut and by reordering clauses. Moreover, there can be no way to make a program consistent by using cut and reorderings. As a consequence, all the time used to build that program is wasted. As an example, suppose we are given the following hypothesis space:

$c'_1$) int(X,Y,Z) :- head(X,A), tail(X,B), int(B,Y,W), cons(A,W,Z).
$c'_2$) int(X,Y,Z) :- null(X), null(Z).
$c'_3$) int(X,Y,Z) :- null(Z).

with the examples:

$e'_1$) int_pos([a],[a],[a]).
$e'_2$) int_pos([a,b],[c],[]).
$e'_3$) int_neg([a],[b],[a]).

Then we can learn the trace $[c'_1, c'_2]$ from $e'_1$ and the trace $[c'_3]$ from $e'_2$. But $[c'_1, c'_2, c'_3]$ covers $e'_3$, and there is no way to make it consistent using cut or by reordering its clauses. In fact, the first partial trace is responsible for this inconsistency, and hence the time used to learn $[c'_3]$ is totally wasted.

Here it is also possible to understand why we need flattened clauses. Consider the following program for *intersection*, which is equivalent to $[c_2, c_4, c_5]$, but with the three clauses unflattened:

$u_2$) int([A|B],Y,W) :- notmember(A,Y), !, int(B,Y,W).
$u_4$) int([A|B],Y,[A|W]) :- !, int(B,Y,W).
$u_5$) int(_,_,[]).

Now, this program covers int_neg([a],[a],[]), i.e. $[u_2, u_4, u_5] \vdash$ int([a],[a],[]). In fact, clause $u_2$ fails on this example because a is a member of [a]. Clause $u_4$ fails because the empty list cannot be matched with [A|W]. But clause $u_5$ succeeds because its arguments match those of the negative example. As a consequence, this program would be rejected by the induction procedure.

The problem is that, if we use unflattened clauses, it may happen that a clause body is not evaluated because an example does not match the head of the clause. As a consequence, possible cuts in that clause are not evaluated and cannot influence the behavior of the entire program. In our example, the cut in clause $u_4$ has no effect because the output argument of int([a],[a],[]) does not match [A|W], and the body of $u_4$ is not evaluated at all. Then $u_5$ is fired and the negative example is covered. In the flattened version, clause $c_4$ fails only when cons(a,[],[]) is reached, but at that point a cut is in force and clause $c_5$ cannot be activated. Note that program $[u_2, u_4, u_5]$ behaves correctly on the query int([a],[a],X), and gives X=[a] as the only output.

## 4.5 Problem 5: Ordering of Literals

Even the relative position of literals and cut in a clause is significant. Consider again the correct program for *intersection* as above ($[c_2, c_4, c_5]$), but with $c_4$ modified by putting the *cons* literal in front of the antecedent:

$c'_4$) int(X,Y,Z) :- cons(A,W,Z), head(X,A), tail(X,B), int(B,Y,W).

Then, there is no way to get a correct program for intersection using this clause. To rule out the negative example int_neg([a],[a],[]) we must put a cut before the *cons* predicate, in order to prevent the activation of $c_5$. But, then, some positive examples are no longer covered, such as int_pos([a],[],[]). In fact, we have a wrong behavior every time clause $c'_4$ is called and fails, since it prevents the activation on $c_5$. In general, this problem cannot be avoided even by reordering clauses: if we put $c_4'$ after $c_2$ and $c_5$, then int_neg([a],[a],[]) will be covered. As a consequence, we should also test for every possible permutation of literals in every clause of a candidate program.

## 5. Situations where Learning Cut is still Practical

From the above analysis, learning cut appears to be difficult since, in general, a learning procedure should be able to backtrack on the candidate base programs (e.g., traces), on the position of cut(s) in the program, on the order of the clauses in the program, on the order of literals in the clauses and on the order of given positive examples. However, we have spotted some general conditions at which learning cut could still be practical. Clearly, these conditions cannot be a final solution to learning cut, but, if applicable, can alleviate the computational problems of the task.

### 5.1 Small Hypothesis Space

First of all, a restricted hypothesis space is necessary. If clauses cannot be learned independently of one another, a small hypothesis space would help to limit the backtracking required on candidate traces (problem 1). Moreover, even the number of clauses in a trace would be probably smaller, and hence also the number of different permutations and the number of different positions for inserted cuts (problems 2 and 1). A small trace would also have a slight positive impact on the need to test for different literal orderings in clauses (problem 5).

In general, many kinds of constraints can be applied to keep a hypothesis space small, such as ij-determinism (Muggleton & Feng, 1990), rule sets and schemata (Kietz & Wrobel, 1991; Bergadano & Gunetti, 1993), determinations (Russell, 1988), locality (Cohen, 1993), etc (in fact, some of these restrictions and others, such as those listed in Section 3, are available in the actual implementation of our procedure - see the Appendix[4]). Moreover, candidate recursive clauses must be designed so that no infinite chains of recursive calls can take place (Bergadano & Gunetti, 1993) (otherwise the learning task itself could be non-terminating). In general, the number of possible recursive calls must be kept small, in order to avoid too much backtracking when searching for possible traces. However, general constraints may not be sufficient. The hypothesis space must be designed carefully from the very beginning, and this can be difficult. In the example of learning $simplify$ an initial hypothesis space of "only" 8449 clauses was obtained specifying not only the set of required predicates, but even the variables occurring in every literal.

If clauses cannot be learned independently, experiments have shown to us that a dramatic improvement of the learning task can be obtained by generating the clauses in the hypothesis space so that recursive clauses, and in general more complex clauses, are taken into consideration after the simpler and non-recursive ones. Since simpler and non recursive clauses require less time to be evaluated, they will have a small impact on the learning time. Moreover, learning simpler clauses (i.e. shorter) also alleviates problem 5.

---

4. We found these constraints particularly useful. By using them we were often able to restrict a hypothesis space of one order of magnitude without ruling out any possible solution.

Finally, it must be noted that our induction procedure does not necessarily require that the hypothesis space S of possible clauses be represented explicitly. The learning task could start with an empty set S and an implicit description of the hypothesis space, for example the one given in Section 3. When a positive example cannot be derived from S, a new clause is asked for to a clause generator and added to S. This step is repeated until the example is derivable from the updated S, and then the learning task can proceed normally.

## 5.2 Simple Examples

Another improvement can be achieved by using examples that are as simple as possible. In fact, each example which may involve a recursive call is potentially responsible for the activation of all the corresponding clauses in the hypothesis space. The more complex the example, the larger the number of consecutive recursive activations of clauses and the larger the number of traces to be considered for backtracking (problem 1). For instance, to learn the *append* relation, it may be sufficient to use an example like append([a],[b],[a,b]) instead of one like append([a,b,c,d],[b],[a,b,c,d,b]). Since simple examples would probably require a smaller number of different clauses to be derived, this would result in smaller traces, alleviating the problem of permutation of clauses and literals in a trace (problems 2 and 5) and decreasing the number of positions for cuts (problem 1).

## 5.3 Small Number of Examples

Since a candidate program is formed by taking the union of partial traces learned for single examples, if we want a small trace (problems 2 and 5) we must use as few examples as possible, while still completely describing the required concept. In other words, we should avoid redundant information. For example, if we want to learn the program for *append*, it will be normally sufficient to use only one of the two positive examples append([a],[b],[a,b]) and append([c],[d],[c,d]). Obviously it may happen that different examples are derived by the same set of clauses, and in this case the final program does not change.

Having to check for all possible orderings of a set of positive examples, a small number of examples is also a solution to problem 4. Fortunately, experiments have shown that normally very few positive examples are needed to learn a program, and hence the corresponding number of different orderings is, in any case, a small number. Moreover, since in our method a positive example is sufficient to learn all the clauses necessary to derive it, most of the time a complete program can be learned using only one well chosen example. If such an example can be found (as in the case of the learning task of section 3, where only one example of *simplify* and one of *remove* are given), the computational problem of testing different example orderings is automatically solved.

However, it must be noted that, in general, a small number of examples may not be sufficient, except for very simple programs. In fact, if we want to learn logic programs such as *member*, *append*, *reverse* and so on, then any example involving recursion will be sufficient. But for more complex programs the choice may not be trivial. For example, our procedure is able to learn the *quicksort* (plus *partition*) program with only one "good" example. But if one does not know how *quicksort* and *partition* work, it is likely that she or he will provide an example allowing to learn only a partial description of *partition*. This is particularly clear in the example of *simplify*. Had we used the positive example

simplify_pos([[[]],[b,a,a]]],[b,a]) (which is very close to the one effectively used), the first clause of *flatten* would not have been learned. In other words, to give few examples we must give good examples, and often this is possible only by having in mind (at least partially and in an informal way) the target program. Moreover, for complex programs, good examples can mean complex examples, and this is in contrast with the previous requirement. For further studies of learning from good examples we refer the reader to the work of Ling (1991) and Aha, Ling, Matwin and Lapointe (1993).

### 5.4 Constrained Positions for Cut and Literals

Experiments have shown that it is not practical to allow the learning procedure to test all possible positions of cut in a trace, even if we are able to keep the number of clauses in a trace small. The user must be able to indicate the positions where a cut is allowed to occur, e.g., at the beginning of a clause body, or before a recursive call. In this case, many alternative programs with cut are automatically ruled out and thus do not have to be tested against the negative examples. It may also be useful to limit the maximum number of cuts per clause or per trace. For example, most of the time one cut per clause can be sufficient to learn a correct program. In the actual implementation of our procedure, it is in fact possible to specify the exact position of cut w.r.t. a literal or a group of literals within each clause of the hypothesis space, when this information is known.

To eliminate the need to test for different ordering of literals (problem 5), we may also impose a particular global order, which must be maintained in every clause of the hypothesis space. However this requires a deep knowledge of the program we want, otherwise some (or even all) solutions will be lost. Moreover, this solution can be in contrast with a use of constrained positions for cut, since a solution program for a particular literal ordering and for particular positions for cuts may not exist.

## 6. Conclusion

Our induction procedure is based on an intensional evaluation of clauses. Since the cut predicate has no declarative meaning, we believe that intensional evaluation of clauses cannot be abandoned, independently of the kind of learning method adopted. This can decrease the performance of the learning task, compared with extensional methods, which examine clauses one at a time without backtracking. However, the computational problems outlined in Section 4 remain even if we choose to learn a complete program extensionally, and then we try to make it consistent by inserting cut. The only difference is that we do not have backtracking (problem 1), but the situation is probably worse, since extensional methods can fail to learn a complete program even if it exists in the hypothesis space. (Bergadano, 1993a).

Even if the ability to learn clauses containing procedural predicates like cut seems to be fundamental to learning "real" logic programs, in particular short and efficient programs, many problems influencing the complexity of the learning task must be faced. These include the number and the relative ordering of clauses and literals in the hypothesis space, the kind and the relative ordering of given examples. Such problems seem to be related to the need for an intensional evaluation of clauses in general, and not to the particular learning method adopted. Even just to alleviate these problems, it seems necessary to know a lot about the target program. An alternative solution is simply to ignore some of the problems. That is, avoid testing for different clause and/or literal and/or example orderings. Clearly, in this way the learning process can become feasible, but it can fail to find a solution even when it exists. However, many ILP systems (such as Foil) adopt such an "incomplete-but-fast" approach, which is guided by heuristic information.

As a consequence, we view results presented in this paper as, at least partially, negative. The problems we raised appear computationally difficult, and suggest that attention should be restricted to purely declarative logic languages, which are, in any case, sufficiently expressive.

## Acknowledgements

This work was in part supported by BRA ESPRIT project 6020 on Inductive Logic Programming.

## Appendix A

The induction procedure of Section 2 is written in C-prolog (interpreted) and runs on a SUNsparcstation 1. We are planning to translate it in QUINTUS prolog. This Appendix contains a simplified description of its implementation. As a preliminary step, in order to record a trace of the clauses deriving a positive example e+, every clause in the hypothesis space[5] S must be numbered and modified by adding to its body two literals. The first one, $allowed(n,m)$ is used to activate only the clauses which must be checked against the negative examples. The second one, $marker(n)$, is used to remember that clause number $n$ has been successfully used while deriving e+. Hence, in general, a clause in the hypothesis space S takes the following form:

$P(X_1,\ldots,X_m) :\text{-} allowed(n,m),\gamma,marker(n).$

where $\gamma$ is the actual body of the clause, $n$ is the number of the clause in the set and $m$ is a number used to deal with cuts. For every clause $n$, the one without cut is augmented with $allowed(n,0)$, while those containing a cut somewhere in their body are augmented with $allowed(n,1)$, $allowed(n,2)$, ..., and so on. Moreover, for every augmented clause as above, a fact "$alt(n,m)$." is inserted in S, in order to implement an enumeration mechanism.

A simplified (but running) version of the learning algorithm is reported below. In the algorithm, the output, if any, is the variable Trace containing the list of the (numbers of the) clauses representing the learned program P. By using the backtracking mechanism of Prolog, more than one solution (trace) can be found. We assume the two predicates *listpositive* and *listnegative* build a list of the given positive and negative examples, respectively.

consult(file_containing_the_set_of_clauses_S).

---

5. We assume clauses in the hypothesis space to be flattened

allowed(X,0).

marker(X) :- assert(trace(X)).
marker(X) :- retract(trace(X)), !, fail.

main :- listpositive(Posexamplelist), tracer([],Posexamplelist,Trace).

tracer(Covered,[Example|Cdr],Trace) :- Example, /* **backtracking point 1** */
                        setof(L,trace(L),Trace1),
                        notneg(Trace1,[Example|Covered],Cdr),
                        tracer([Example|Covered],Cdr,Trace).
tracer(_,[],Trace) :- setof((I,J),allowed(I,J),Trace), asserta((marker(X) :- true, !)).

assertem([]).
assertem([I|Cdr]) :- alt(I,J), backassert(allowed(I,J)), assertem(Cdr).

prep(T) :- retract(allowed(X,0)), assertem(T).

backassert(X) :- assert(X).
backassert(X) :- retract(X), !, fail.

resetallowed([]) :- !.
resetallowed(_) :- abolish(allowed,2), assert(allowed(X,0)), !.

notneg(T,Covered,Remaining) :- listnegative([]).
notneg(T,Covered,Remaining) :- listnegative(Negexamplelist),
                        asserta((marker(X) :- true,!)),
                        prep(T), /* **backtracking point 2** */
                        trypos(Covered), trynegs(Negexamplelist),
                        resetallowed(Remaining),
                        retract((marker(X) :- true,!)).
notneg(T,Covered,Remaining) :- resetallowed(Remaining),
                        retract((marker(X) :- true,!)), !, fail.

trypos([Example|Cdr]) :- Example, !, trypos(Cdr).
trypos([]) :- !.

trynegs([Example|Cdr]) :- Example,!,fail.
trynegs([Example|Cdr]) :- trynegs(Cdr).
trynegs([]) :- !.

Actually, our complete implementation is more complex, also in order to achieve greater
efficiency. The behavior of the learning task is quite simple. Initially, the set S of clauses is
read into the Prolog interpreter, together with the learning algorithm. Then the learning
task can be started by calling the predicate *main*. A list of the positive examples is formed and the *tracer* procedure is called on that list. For every positive example, *tracer* calls the example itself, firing all the clauses in S that may be resolved against that example. Observe that, initially, an *allowed*(X,0) predicate is asserted in the database: in this way only clauses not containing a cut are *allowed* to be used (this is because clauses with cut are employed only if some negative example is derived). Then, a trace, if any, of (the numbers associated to) the clauses successfully used in the derivation of that example is built, using the *setof* predicate.

The trace is added to the traces found for the previous examples, and the result is checked against the set of the negative examples calling the *notneg* procedure. If *notneg* does not fail (i.e. no negative examples are covered by this trace) then a new positive example is taken into consideration. Otherwise *notneg* modifies the trace with cut and tests it again. If also this fails, backtracking occurs and a new trace for the current example (and possibly for the previous ones) is searched for.

The *notneg* procedure works as follows. First, only the clauses in the trace are *allowed* to be checked against the negative examples, by retracting the *allowed*(X,0) clause and asserting an *allowed*($n$,0) if the $n$-th clause (without cut) is in the trace. This is done with the *prep* and *assertem* predicates. Then a list of the negative examples is formed and we check if they can be derived from the clauses in the trace. If at least one negative example is covered, (i.e., if *trynegs* fails) then we backtrack to the *prep* procedure (backtracking point 2) where a clause of the trace is substituted with an equivalent one but with cut inserted somewhere (or in a different position). If no correct program can be found in such a way by trying all possible alternatives (i.e. by using cut in all possible ways), *notneg* fails, and backtracking to backtracking point 1 occurs, where another trace is searched for. Otherwise, all clauses in S without cut are reactivated by asserting again *allowed*(X,0), and the next positive example is considered. Note that *trypos* is used in *notneg* to verify if a modified trace still derives the set of positive examples derived initially. The possibility to substitute clauses in the current trace with others having cut inserted somewhere is achieved through the *alt* predicate in the *assertem* procedure. Finally, note that this simplified version of the learning procedure is not able to generate and test for different orderings of clauses in a trace or for different ordering of literals in each clause, nor to use different orderings for the set of positive examples.

In order to derive all the positive examples before the check against the negative ones (see subsection 4.4), we must change the first clause of the *tracer* procedure into:

tracer([Pos1, ... ,Posn]):-Pos1, ... ,Posn, setof(L,trace(L),T), notneg(T).

The actual implementation of the above induction procedure is available through ftp. For further information contact gunetti@di.unito.it.